# Neuromorphic Online Learning for Spatiotemporal Patterns with a Forward-only Timeline

Zhenhang Zhang, Jingang Jin, Haowen Fang, Qinru Qiu

Department of Electrical Engineering & Computer Science, Syracuse University

{ zzhan281, jjin24, hfang02}@syr.edu, qinru.qiu@gmail.com

*Abstract*—Spiking neural networks (SNNs) are bio-plausible computing models with high energy efficiency. The temporal dynamics of neurons and synapses enable them to detect temporal patterns and generate sequences. While Backpropagation Through Time (BPTT) is traditionally used to train SNNs, it is not suitable for online learning of embedded applications due to its high computation and memory cost as well as extended latency. Previous works have proposed online learning algorithms, but they often utilize highly simplified spiking neuron models without synaptic dynamics and reset feedback, resulting in subpar performance. In this work, we present Spatiotemporal Online Learning for Synaptic Adaptation (SOLSA), specifically designed for online learning of SNNs composed of Leaky Integrate and Fire (LIF) neurons with exponentially decayed synapses and soft reset. The algorithm not only learns the synaptic weight but also adapts the temporal filters associated to the synapses. Compared to the BPTT algorithm, SOLSA has much lower memory requirement and achieves a more balanced temporal workload distribution. Moreover, SOLSA incorporates enhancement techniques such as scheduled weight update, early stop training and adaptive synapse filter, which speed up the convergence and enhance the learning performance. When compared to other non-BPTT based SNN learning, SOLSA demonstrates an average learning accuracy improvement of 14.2%. Furthermore, compared to BPTT, SOLSA achieves a 5% higher average learning accuracy with a 72% reduction in memory cost.

## I. Introduction

Neuromorphic computing offers a promising alternative to conventional artificial neural networks, primarily due to its remarkable energy efficiency. Inspired by the neural system in the human brain, neuromorphic computing builds machine intelligence on top of spiking neural networks (SNN) [23][24], where information is represented through sequences of spiking activities. Due to this nature, the operations of SNN are event driven, in other words, neurons communicate and compute only when there are output or input activities. This event driven approach, combined with closely integrated memory and computation, results in reduced workload and improved energy efficiency. These properties are frequently harnessed by cutting-edge neuromorphic hardware like Intel Loihi [1] and IBM TrueNorth [2], to achieve substantial energy reduction.

Unlike conventional artificial neural networks (ANNs), SNNs have the unique ability to retain past information within the membrane potential of their neurons and synapses. This membrane potential is considered as the state of neurons and synapses. The changes of the membrane potential over time, which are influenced not only by current inputs but also by historical inputs, is referred to as temporal dynamics. Many SNN models take into account the temporal dynamics in the membrane potential of neurons. For example, the widely used Leaky Integrate and Fire (LIF) neuron [3], model membrane potential as a leaky integration of the input with exponential decay. Once the membrane potential reaches a certain threshold, the neuron fires a spike and resets its membrane potential. In a more advanced model proposed by [4], called spatial temporal (ST) SNN, the temporal dynamics of synapses are also considered. It has been demonstrated both biologically [5] and computationally [6] that the temporal dynamics in neurons and synapses play a crucial role in enabling the network to perform spatiotemporal operations effectively.

A basic method of SNN training is to update the weights based on the timing of pre- and post-synaptic neuron spiking events, which is referred to as Spike-Timing-Dependent-Plasticity (STDP) rule [7]. When a pre-synaptic spiking event happened before post-synaptic spiking event within a time window, this leads to a potentiation of the synaptic weight. A contrast condition leads to a depression of weight. This weight update rule is also called Hebbian Learning Rule [8]. Vanilla Hebbian Rule is usually slow and less effective compared to the backpropagation algorithm used in ANN training [9].

By modeling the temporal dynamics of neurons and synapses as filters with feedback connections [4], an SNN can be seen as a recurrent neural network (RNN) that can be trained using Backpropagation Through Time (BPTT) [10][21][22]. This algorithm unrolls the network along the temporal axis and applies conventional backpropagation to the unrolled network. The gradients need to propagate from the output layer all the way back to the input layer and from present time back to the time when the input sequence begins. Therefore, BPTT requires a significant amount of memory to store the historical activities of neurons. Additionally, the learning process cannot start until the entire temporal sequence has been received. This either leads to an extended processing time or a high workload surge at the end of each input sequence. These limitations make BPTT unsuitable for embedded systems, such as personalized health monitors, sensors, and actuators in autonomous vehicles, where online learning of spatiotemporal patterns is required for the system to adapt to a specific user or a new application environment.

The limitations discussed above regarding BPTT have spurred the development of online training algorithms for SNN, such as eligibility trace propagation (i.e., E-prop) [11], Online Training Through Time (OTTT) [12] and Deep Continuous Local Learning (DECOLLE) [13]. However, these algorithms ignore the temporal dynamics of the synapses [11][12]. Furthermore, some of these algorithms [12] overlook the reset process of membrane potential. Consequently, their performance is compromised, particularly when the input sequence gets longer.

In this work we present SOLSA (Spatiotemporal Online Learning for Synaptic Adaptation), a learning algorithm to train SNNs with temporal filters on both synapses and neuron membrane potentials. SOLSA goes beyond learning synaptic weights and can update synapse filter kernels as well. Moreover, it incorporates techniques like scheduled weight update and early stop, which further accelerates the convergence and enhances model accuracy. As a result, SNN models trained using SOLSA achieve an average improvement of 14.2% in spatiotemporal pattern classification compared to models trained using existing online learning algorithms. Furthermore, SOLSA outperforms BPTT by improving the classification accuracy by 5% while requiring 72% less memory.

The rest of the paper is organized as the following. In Section II, we will introduce related works. Section III will present details of our proposed learning algorithm. In Section IV, we will show experimental results of SOLSA on different spatial temporal datasets and compare it with other learning algorithms. Section V gives the conclusions.

## II. BACKGROUND AND RELATED WORKS

The temporal dynamics of a spiking neuron comes from the temporal filters associated to its membrane potentials. The neuron membrane potential of a LIF neuron is the leaky integration of its input, which can be updated incrementally as the following equation:

$$V[t] = \lambda V[t-1] + \sum_j^N w_j I_j[t] - V_{th} O[t-1]$$

where $V[t]$ stands for the membrane potential of the neuron at time $t$, $w_j$ is the weight of the $j$th input of the neuron. $O[t]$ stands for the output of the neuron. It is either 0 or 1 and calculated using the Heaviside activation function $U(\ )$ as the following expression,

$$O[t] = U(V[t] - V_{th}) \begin{cases} 0, V[t] \leq V_{th} \\ 1, otherwise \end{cases}, \quad (1)$$

where $V_{th}$ is a threshold. Fig. 1 illustrates the dataflow of a LIF neuron model.

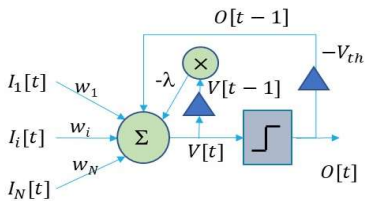

Fig. 1. LIF neuron has a recurrent network structure.

### A. BPTT Based SNN Training

As shown in Fig. 1, the LIF neuron has a recurrent structure. When a network is formed by connected LIF neurons, it becomes a recurrent neural network that can be trained using backpropagation through time (BPTT) [25][26]. Since the Heaviside activation function of spiking neurons is not differentiable, a surrogate function is commonly used as an approximation of gradients. In [14], different types of surrogate functions were numerically studied to understand their impact on SNN learning performance.

Despite being an effective learning algorithm, BPTT has a significant drawback that prohibits its implementation on edge device for online learning [27]. Before initiating backpropagation (i.e., learning), the entire input sequence, which can be quite long, must be received and complete the forward propagation process. This not only increases the latency but also causes an unbalanced temporal distribution of workload. Additionally, the delayed learning requires all neurons and synapses maintain a cycle-accurate record of their activities during the processing of input data, as this information is needed for backpropagation at the end. Consequently, a large memory is needed.

Several techniques have been proposed to improve the performance of BPTT-based SNN learning by taking advantage of the sparsity and binary nature of SNN [15] [16]. For example, [15] proposes a spiking ResNet that represents the residual signal and the output of convolutional layer as spike trains. This enables the implementation of the element wise residual function using a simple AND gate. Such modification simplifies the computation and addresses the issue with vanishing/exploding gradient problem during learning. In [16], a sparse SNN backpropagation algorithm is proposed in. This algorithm leverages the sparsity of SNN in the gradient backpropagation. Neurons are considered active only when their potential is close to the threshold. During the BPTT process, the algorithm calculates the spike gradients solely for active neurons, setting the gradient to zero otherwise. As a result, this approach reduces computation complexity. However, it is important to note that the complexity reduction of above techniques is sporadic, and their worst-case complexity remains unchanged. As BPTT based algorithms, they inherit all the limitations mentioned earlier, making them unsuitable for online learning.

To address the limitations of the original BPTT and enable edge applications, Truncated BPTT was proposed [17]. It truncates the original sequence to a bounded history considering only a fixed number of time steps during backpropagation. However, as a variation of the BPTT algorithm, truncated BPTT still requires maintaining a record of history for certain time steps. Therefore, it does not fundamentally change the overall memory and computation complexity. Furthermore, by ignoring long term history, Truncated BPTT introduces bias. Consequently, its performance deteriorates when dealing with very long sequences.

### B. Non-BPTT Based SNN Training

To enable online learning and edge implementation, recent works have employed three-factor Hebbian Learning [18][19]. This type of learning typically incorporates three factors in the weight update process: a term that represents the presynaptic activities, a term that represents the postsynaptic activities, and a term that represents the neuromodulated global error signal. Additionally, the first two factors can be combined to form an eligibility trace [18], which serves as a transient memory or a flag. In a biological system, the eligibility trace is activated when there is a coincidence of spikes between a presynaptic neuron and a postsynaptic neuron. When the eligibility trace is active, signals, which may indicate novelty or surprise, can impact the connection weights between neurons.

Several three-factor Hebbian learning algorithms have been proposed for online learning in SNNs [11][12][13]. One of them is Online Training Through Time (OTTT) [12], which trains SNNs with LIF neurons. OTTT avoided propagating gradients backward through time by disregarding the dependency of the neuron membrane potential on the output activities in the previous time step. In other words, it ignores the membrane potential reset process. While this approximation simplifies the backpropagation process, it also leads to a degradation in learning performance. Another notable work [11], known as E-prop learning, calculates weight change as an accumulation of the product of two variables, the eligibility trace and the learning signal. The former is incrementally computed in a forward manner, while the latter approximates the influence of the spike output of a neuron on output error. Reference [13] proposed a method call DECOLLE which attaches random readout matrices to each layer and defines the global loss as sum of each layer's loss. DECOLLE set all non-local gradients as zero to enforce locality. The weight update rule can then be defined as three parts, errors, pre-synaptic activity, and post-synaptic activity. The weights will be updated at each time step based on three parts in update rule. The algorithms based on three-factor Hebbian learning calculate weight changes incrementally without unrolling the network, making them suitable for online learning and edge implementation. However, existing algorithms of this kind tend to overlook the temporal dynamics on the synapses, which results in a degradation of learning performance as the sequence lengthens.

The proposed SOLSA learning can also be classified as a Three-Factor Hebbian Learning approach. It tackles the limitations of BPTT by incrementally updating the gradient through forward time-axis updates and interleaving forward and backward propagation. In comparison to the BPTT algorithm, SOLSA offers benefits such as reduced memory usage, shorter latency, and a more balanced distribution of workload. As a result, it is better suited for online training in edge applications. Furthermore, SOLSA distinguishes itself from other three-factor Hebbian learning algorithms by considering synapse temporal dynamics and incorporating various enhancement techniques. These include adaptive filter kernel, scheduled weight updates, and early stop mechanisms, all of which significantly contribute to the improved training performance of SOLSA.

## III. PROPOSED METHOD

In this section, we will begin by introducing the neuron and synapse model that SOLSA considers. We will then provide detailed information about the learning algorithm, which includes several enhancement techniques aimed at improving learning performance. A summary of the notations used in this paper is given in Table 1.

### A. Neuron and Synapse Models

In this work, we follow [4] to consider a layered network structure with temporal dynamics in both neurons and synapses. The membrane potential $V_i^l[t]$ of the $i$th neuron at layer $l$ at time $t$ is updated based on the following equation:

$$V_i^l[t] = \lambda V_i^l[t-1] + \sum_j^{N_{l-1}} w_{i,j}^l F_{i,j}^l[t] - V_{th} O_i^l[t-1], \quad (2)$$

where $O_i^l[t-1]$ denotes the output of this neuron in previous time step. The output is calculated using the Heaviside activation function given in Equation (1), $O_i^l[t] = U(V_i^l[t] - V_{th})$, where $V_{th}$ is potential threshold, $w_{i,j}^l$ is the weight coefficient and $F_j^l[t]$ is the post synaptic potential of the $j$th input. To model the filter effect of synapses, $F_j^l[t]$ is the output of a first order IIR filter updated as the follows:

$$F_j^l[t] = \alpha_{ij}^l F_j^l[t-1] + \beta_{ij}^l O_j^{l-1}[t-1] \quad (3)$$

Equation (3) shows that the output of synapse filter in current time step is determined by its previous value and the input (i.e., $O_j^{l-1}[t-1]$) in previous time stamps. $\alpha_{ij}^l$ and $\beta_{ij}^l$ are coefficients of the filter that control the balance between the forward and backward taps of the filter.

Fig. 2 depicts a diagram of the neuron model. Based on the model, the neuron membrane potential is a leaky integration of the synaptic potential. If a neuron fired in previous time step, its potential $V_i^l[t]$ should decrease by the threshold value as a reset.

Fig. 3 illustrates the data flow graph of the SNN after being unrolled over time. Each column represents a single time step. The duration of the time step is determined by the intervals between consecutive data in the input sequence. During inference, the signal propagates from bottom to top and left to right. During training, when applying BPTT, the gradient propagates reversely from top right corner to the bottom left corner. The vertical propagation is referred to as the *spatial propagation* and the horizontal propagation is referred to as the *temporal propagation*. All parameters in the SNN, including $w_{ij}^l$, $\alpha_{ij}^l$, and $\beta_{ij}^l$, can be trained using BPTT.

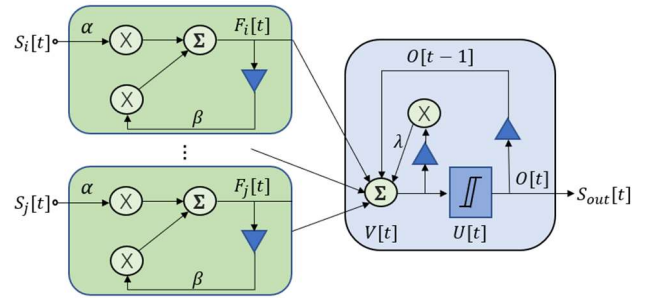

Fig. 2. The architecture of the neuron model.

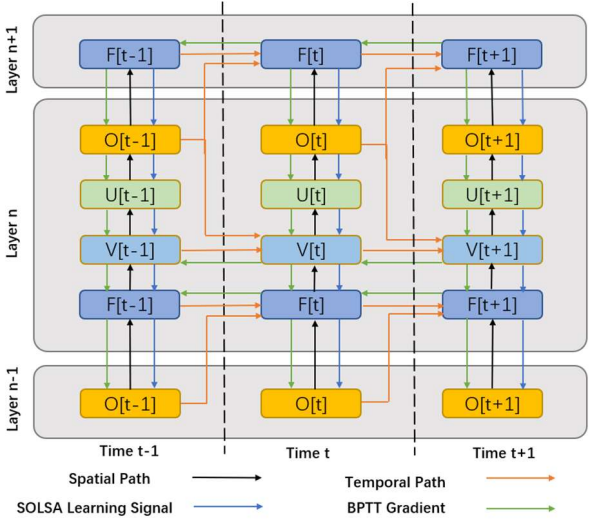

Fig. 3. Unrolled ST SNN.

Table 1 Notations Used in the Paper

| Symbol | Description |
|---|---|
| $V_i^l[t]$ | Membrane potential of the $i$th neuron in layer $l$ at time $t$ |
| $O_i^l[t]$ | Output of the $i$th neuron in layer $l$ at time $t$ |
| $F_{ij}^l[t]$ | Synapse potential of the $ij$th connection in layer $l$ (i.e., connection from the $j$th neuron in layer $l-1$ to the $i$th neuron in layer $l$) |
| $\alpha_{ij}^l, \beta_{ij}^l$ | Coefficients of synaptic filter of $ij$th connection in layer $l$ |
| $w_{i,j}^l$ | Synaptic weight of $ij$th connection in layer $l$ |
| $U(\cdot)$ | Heaviside activation function of spiking neurons |
| $\epsilon_i^l[t]$ | Surrogate gradient of the Heaviside activation function |
| $\mu_i^l[t]$ | Gradient of $V_i^l[t]$ backpropagated through the spatial path |
| $E$ | Output error (difference of the actual and expected output) |
| $E[t]$ | Output error at time $t$ |

*B. SOLSA Learning*

The Heaviside activation function is non-differentiable [28]. However, under a Gaussian noise $z \sim N(0, \sigma)$, the probability that a neuron with membrane potential $V$ and threshold $V_{th}$ will fire an output spike, which can be calculated using Equation (4),

$$P(V + z > V_{th}) = \frac{1}{2}\text{erfc}\left(\frac{V_{th} - V}{\sqrt{2}\sigma}\right), \quad (4)$$

is differentiable. Following many previous works [4][14], we adopt the derivative of the spiking probability as the surrogate gradient for the Heaviside activation. Let $\epsilon_i^l[t]$ denote the derivative of spike activity of neuron $i$ in layer $l$, $\epsilon_i^l[t] \approx \frac{dP(V_i^l[t]+z>V_{th})}{dV_i^l[t]}$.

We denote $E$ as the error, which is the difference between the actual output and the expected output. The term "$ij$th connection in layer $l$" refers to the connection from the $j$th neuron in layer $l-1$ to the $i$th neuron in layer $l$. With the given error, the gradient of the weight $w_{ij}^l$ of the $ij$th connection in layer $l$ can be expressed as Equation (5):

$$\frac{dE}{dw_{ij}^l} = \sum_t \frac{dE}{dV_i^l[t]} \frac{\partial V_i^l[t]}{\partial w_{ij}^l}, \quad (5)$$

where the term $\frac{dE}{dV_i^l[t]}$ calculates the impact of the membrane potential $V_i^l[t]$ on the error $E$.

The $V_i^l[t]$ affects the error in two ways. Firstly, through the spatial path, it determines the neuron's output $O_i^l[t]$, which subsequently propagates to neurons in layer $l+1$ and beyond, ultimately reaching the output layer to determine the output error in the current time step. Secondly, through the temporal path, $V_j^l[t]$ determines the neuron's membrane potential in the next time step (i.e., $V_j^l[t+1]$) via the leaky integrate process, which will further impact the future output error. Therefore, we can rewrite Equation (5) as the following expression:

$$\frac{dE}{dV_i^l[t]} = \underbrace{\frac{dE}{dO_i^l[t]} \frac{\partial O_i^l[t]}{\partial V_i^l[t]}}_{\text{Spatial Path}} + \underbrace{\frac{dE}{dV_i^l[t+1]} \frac{\partial V_i^l[t+1]}{\partial V_i^l[t]}}_{\text{Temporal Path}}, \quad (6)$$

First, we analyze the spatial path. Using the surrogate gradient function, we know that $\frac{\partial O_i^l[t]}{\partial V_i^l[t]} = \epsilon_i^l[t]$. Using the chain rule, the spatial gradient can be further expanded as follows:

$$\frac{dE}{dO_i^l[t]} \frac{\partial O_i^l[t]}{\partial V_i^l[t]} \approx \sum_k \frac{\partial E[t]}{\partial O_k^{l+1}[t]} \frac{\partial O_k^{l+1}[t]}{\partial V_k^{l+1}[t]} \frac{\partial V_k^{l+1}[t]}{\partial F_{ik}^{l+1}[t]} \frac{\partial F_{ik}^{l+1}[t]}{\partial O_i^l[t]} \frac{\partial O_i^l[t]}{\partial V_i^l[t]}$$

$$= \sum_k \frac{\partial E[t]}{\partial O_k^{l+1}[t]} \epsilon_k^{l+1}[t] w_{ki}^{l+1} \beta_{ki}^{l+1} \epsilon_i^l[t]. \quad (7)$$

Please note that we have made certain approximations in this step by omitting the temporal path for neurons in the $(l+1)$th layer and above. Some of this temporal impact will be taken into account in next time step when we calculate $\frac{dE}{dV_i^l[t+1]}$. As we can see, Equation (7) calculates the vertical backpropagation from the output layer to the $l$th layer. For the simplicity, we denote the spatial path gradient in (7) as $\mu_i^l[t]$ and refer to it as the *learning signal*.

Next, we analyze the temporal path in Equation (6). Note that $\frac{dE}{dV_i^l[t+1]}$ is just the one step shifted version of $\frac{dE}{dV_i^l[t]}$, we can rewrite (6) as the following:

$$\frac{dE}{dV_i^l[t]} = \mu_i^l[t] + \left(\mu_i^l[t+1] + \frac{dE}{dV_i^l[t+2]} \frac{\partial V_i^l[t+2]}{\partial V_i^l[t+1]}\right) \frac{\partial V_i^l[t+1]}{\partial V_i^l[t]} \quad (8)$$

By recursively decomposing $\frac{dE}{dV_i^l[t+2]}$, we rewrite Equation (8) as the follows:

$$\frac{dE}{dV_i^l[t]} = \sum_{t \leq t' \leq T} \mu_i^l[t'] \frac{\partial V_i^l[t']}{\partial V_i^l[t'-1]} \frac{\partial V_i^l[t'-1]}{\partial V_i^l[t'-2]} \cdots \frac{\partial V_i^l[t+1]}{\partial V_i^l[t]}, \quad (9)$$

where $T$ is the length of the entire input sequence. Plugging (9) into (5) we get (10).

$$\frac{dE}{dw_{ij}^l} = \sum_t \sum_{t \leq t' \leq T} \mu_i^l[t'] \frac{\partial V_i^l[t']}{\partial V_i^l[t'-1]} \frac{\partial V_i^l[t'-1]}{\partial V_i^l[t'-2]} \cdots \frac{\partial V_i^l[t+1]}{\partial V_i^l[t]} \frac{\partial V_i^l[t]}{\partial w_{ij}^l}$$

$$= \sum_{t' \leq T} \mu_i^l[t'] \cdot \underbrace{\sum_{t \leq t'} \frac{\partial V_i^l[t']}{\partial V_i^l[t'-1]} \frac{\partial V_i^l[t'-1]}{\partial V_i^l[t'-2]} \cdots \frac{\partial V_i^l[t+1]}{\partial V_i^l[t]} \frac{\partial V_i^l[t]}{\partial w_{ij}^l}}_{\varepsilon_{i,j}^l[t']} \quad (10)$$

We denote $\sum_{t \leq t'} \frac{\partial V_i^l[t']}{\partial V_i^l[t'-1]} \frac{\partial V_i^l[t'-1]}{\partial V_i^l[t'-2]} \cdots \frac{\partial V_i^l[t+1]}{\partial V_i^l[t]} \frac{\partial V_i^l[t]}{\partial w_{ij}^l}$ using $\varepsilon_{i,j}^l[t']$ and refer to it as the *Eligibility Trace*. The Eligibility Trace can be updated incrementally using (11):

$$\varepsilon_{i,j}^l[t'] = \varepsilon_{i,j}^l[t'-1] \frac{\partial V_j^l[t']}{\partial V_j^l[t'-1]} + \frac{\partial V_i^l[t']}{\partial w_{ij}^l}$$

$$= (\lambda - v_{th}\epsilon_i^l[t'])\varepsilon_{i,j}^l[t'-1] + F_{i,j}^l[t'] \qquad (11)$$

Equation (11) shows that the eligibility trace can be implemented as a leaky integration of the synaptic potential $F_{i,j}^l[t]$, with leakage coefficient $(\lambda - V_{th}\epsilon_i^l[t])$.

Using the learning signal ($\mu_j^{l+1}[t']$) and the eligibility trace $\varepsilon_{i,j}^l[t']$, we now have the SOLSA learning rule.

$$\frac{dE}{dw_{ij}^l} = \sum_{t'} \mu_j^{l+1}[t'] \cdot \varepsilon_{i,j}^l[t'] \qquad (12)$$

Both the learning signal and eligibility trace rely only on signals of current time or previous time. The entire algorithm can be implemented using a streaming process. It always moves forward in time.

Fig. 4 illustrates how the computing workload of BPTT [29] and SOLSA is distributed over time. For applications with streamed input, the system receives input data periodically at a given sampling rate. Each input data undergoes forward propagation, where the data propagates from the bottom to the top following the spatial path shown in Fig. 3. In SOLSA, spatial backpropagation is then performed to calculate $\mu_j^l[t']$ in each layer, followed by an update of $\varepsilon_{i,j}^l[t']$. In BPTT, the learning does not start until the last data of the input sequence is received, after which the gradients are propagated through the spatial and temporal backpropagation path. In Fig. 4, the size of each block roughly represents the corresponding task's workload. Generally, the computation of the spatial and temporal backpropagation for BPTT is more than the sum of the computations of all spatial backpropagations that occur in each time step in SOLSA. As the input sequence length increases, the computation of backpropagation for BPTT in the last step drastically increases. It is evident that the workload distribution of the BPTT has significant variations over time, while the workload of SOLSA is more evenly distributed. The balanced workload distribution facilitates better energy and latency management for embedded real-time systems.

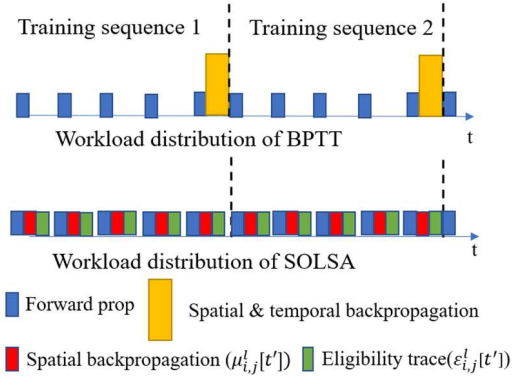

Fig. 4. Compared to BPTT, SOLSA has more balanced workload.

*C. Adaptive Synapse Filter Kernel*

Equation (12) calculates the gradient with respect to the weight coefficients specifically. In our experiment, we noticed that the initial values of $\alpha_{ij}^l$ and $\beta_{ij}^l$ significantly influenced the learning performance. Previous works on BPTT training have also shown that optimizing the filter kernels, achieved by adapting $\alpha_{ij}^l$ and $\beta_{ij}^l$, leads to improved learning results [4]. Therefore, SOLSA also introduces an online technique that approximates the gradient with respect to the filter kernels.

If we only consider time step $t$, the gradient of error $E[t]$ with respect to $\alpha_{ij}^l$ and $\beta_{ij}^l$ can be calculated using the spatial backpropagation path as follows:

$$\nabla_{\alpha_{ij}^l}[t] = \frac{\partial E[t]}{\partial \alpha_{ij}^l} = \mu_i^l[t] \frac{\partial V_i^l[t]}{\partial F_{ij}^l[t]} \frac{\partial F_i^l[t]}{\partial \alpha_{ij}^l} = \mu_i^l[t] \cdot w_{ij}^l \cdot F_i^l[t-1],$$

$$\nabla_{\beta_{ij}^l}[t] = \frac{\partial E[t]}{\partial \beta_{ij}^l} = \mu_i^l[t] \frac{\partial V_i^l[t]}{\partial F_{ij}^l[t]} \frac{\partial F_i^l[t]}{\partial \beta_{ij}^l} = \mu_i^l[t] \cdot w_{ij}^l \cdot O_j^{l-1}[t-1].$$

However, the error $E[t]$ is not solely influenced by neuron activities in time step $t$, but also by activities during past time steps. We assume that the impact on the future error decays exponentially with time. To capture this historical impact, we calculate the overall gradient of $E[t]$ as follows:

$$\frac{dE[t]}{d\beta_{ij}^l} \approx \nabla_{\beta_{ij}^l}[t] \cdot \sum_{n=0}^{t} \gamma^n,$$

$$\frac{dE[t]}{d\beta_{ij}^l} \approx \nabla_{\beta_{ij}^l}[t] \cdot \sum_{n=0}^{t} \gamma^n,$$

where $0 < \gamma < 1$ is a decay factor. Finally, we consider all $E[t]$ from time 0 to T, and calculate the gradient of $\alpha_{ij}^l$ and $\beta_{ij}^l$ using (13) and (14):

$$\frac{dE}{d\alpha_{ij}^l} = \sum_t \nabla_{\alpha_{ij}^l}[t] \cdot \frac{1-\gamma^{t+1}}{1-\gamma}, \qquad (13)$$

$$\frac{dE}{d\beta_{ij}^l} = \sum_t \nabla_{\beta_{ij}^l}[t] \cdot \frac{1-\gamma^{t+1}}{1-\gamma}. \qquad (14)$$

*D. Scheduled weight update*

Unlike BPTT, where gradients and weight updates are calculated at the end of the input sequence, SOLSA learning accumulates weight updates overtime, as shown in (12). This enables us to update the network from time to time to reflect the partial results of learning even if the entire input sequence has not been received. This facilitates faster convergence by allowing the SNN to escape inferior settings more quickly. However, updating weights frequently at each time step can introduce noise due to local variations in the input sequence. Therefore, careful selection of when to update is crucial. We propose an approach called *scheduled weight update*, which selects specific time steps as update points and only update weights at these selected points. The number of update points serves as a hyper-parameter and is configured before training. Generally, a ratio of 1 to 50 between the update points and the sequence length yields desirable results.

We utilize the first several epochs in SOLSA learning to establish an update schedule. Initially, the update schedule consists of only one update point, which is the end of the input sequence. At each time step $t$, we calculate and record the sum of absolute partial gradient, denoted as $g_t$:

$$g_t = \sum_{ijl} \left| \frac{dE[t]}{dw_{ij}^l} \right| = \sum_{ijl} \left| \sum_{t'=0}^{t} \mu_j^l[t'] \cdot \varepsilon_{i,j}^l[t'] \right|.$$

The time steps with the highest $g_t$ values are selected as update points and added to the update schedule. The process of update point selection is summarized as Algorithm 1 depicted in Fig. 5. After the update schedule X is created, it will be used for the rest of the training.

```
Algorithm 1 Flow of update points selection
Require: f(x): model; x_0: initial update schedule {i_end}, i_end equals to the
    length of the input sequence; N: the number of update points;
Ensure: a set of effective update schedule X;
 1: X = {i_end};
 2: repeat
 3:   use x_0 as the update schedule and apply SOLSA learning;
 4:   record the sum of absolute gradient g_t at each time step;
 5:   g = {g_0, g_1, ..., g_t};
 6:   pick the top N largest gradient from g and get their index i_0, i_1, ..., i_{n-1};
 7:   x_0 = {i_0, i_1, ..., i_{n-1}, i_end};
 8:   pick the largest gradient from g and get its index i_max;
 9:   insert the index i_max into the update schedule: X ← i_max;
10: until (|X| > N)
```

Fig. 5. Flow of determining update points.

### E. Eearly-Stop Training

For certain input sequences, the distinctive pattern that signifies the class information emerges right at the beginning. This enables the SNN model to make accurate predictions before reaching the end of the sequence. However, the data that follows the signature pattern may consist of random noise that does not contribute to the classification. Training the model using the data beyond the signature pattern may deteriorate the learned features. To address this issue, we propose an early-stop approach that focuses on the segment of the sequences containing useful information.

For each training sample, the early-stop method involves monitoring the prediction accuracy, denoted as $A(t)$, at every scheduled update point $t$. If the accuracy surpasses a predefined threshold indicating a correct prediction, a counter $C$ is incremented. We will stop processing this training sample when $C$ reaches $N/2$, where $N$ is the total number of update points.

Fig. 6 shows an illustrative example of the early stop training technique. The model performs a binary classification of input sequences. It has two output neurons, which generate spike trains $O_0[t]$ and $O_1[t]$. The output neuron with the higher spiking activities indicates the prediction. The accuracy function is defined as $A(t) = \frac{\sum_{i=0}^{t} O_{target}[i]}{\sum_k \sum_{i=0}^{t} O_k[i]}$, and the threshold is set to be 0.5. In other words, the prediction is considered correct if the neuron corresponding to the target class generates more output spikes than the other neuron. The figure shows the activities of the two output neurons for an input sequence with length $T$. It also shows the accuracy function and the counter value at different update points. The $C$ value reached $N/2$ at the 3rd update point, which indicates an early stop of the training process of this input sample.

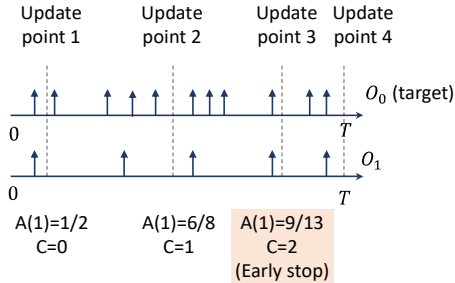

Fig. 6. An example of early-stop training.

As we can see from the example, to trigger an early stop, we need two conditions, an early emergence of the signature pattern and a well-trained model. The model must consistently give correct prediction over time for a given training sequence to early stop the training process. We will show in the experimental results, early stop training not only helps us to dynamically reduce the computation load, but also improves the learning performance. This enhancement technique is available only to non-BPTT based learning, as the BPTT algorithm must wait until the end of the input sequence to start calculating the gradient and to update the model.

## IV. EXPERIMENTAL RESULTS

We tested SOLSA learning on multivariate time series of varying lengths. Sequences with more than 800 time-steps are considered as long sequences, while those with fewer time steps were categorized as regular sequences. TABLE I. summarizes the key statistics of those datasets [31][32][33][34][35][36][37][38][39] including their dimension, length, and input style. Most of the sequences consist of recorded sensor readings. The data are floating point numbers representing the output current of the sensors. To convert these currents into spikes, current-based LIF neurons [30] are employed in the first layer for temporal population coding. All datasets are processed by fully connected SNNs with neuron models specified in Section III.A. The network structures are given on the table. All layers, including the input layer, are trainable.

TABLE I. DATASET INFORMATION

| | Dataset Name | Input size | Sequence length | Input format | SNN architecture |
|---|---|---|---|---|---|
| Regular | EMG gesture | 8 | 100 | Current | 8-150-150-7 |
| | Finger mov. | 28 | 50 | Current | 28-100-100-2 |
| | Basic motion | 6 | 100 | Current | 6-100-100-4 |
| | Epilepsy | 3 | 207 | Current | 3-100-100-4 |
| | Jap. Vowel | 12 | 29 | Current | 12-100-100-9 |
| | RacketSports | 6 | 30 | Current | 6-100-100-4 |
| | DVS128 | 4096 | 100 | Spikes | 4096-100-100-11 |
| Long | Self reg. scp | 6 | 896 | Current | 6-100-100-2 |
| | EMG action | 8 | 1000 | Current | 8-200-200-10 |

Four reference algorithms, namely BPTT [10], Truncated BPTT (T-BPTT) [17], E-prop [11], and OTTT [12], were implemented for comparison. The T-BPTT considers up to 20 steps of historical information during training and disregards past information beyond that point. Both BPTT and T-BPTT require additional storage for the unrolled network, hence are unsuitable for edge implementation. E-prop learning and OTTT do not use synapse filters or other enhancement techniques such as scheduled weight update and early stop, as SOLSA does. Furthermore, OTTT omits the impact of membrane potential reset in the gradient approximation. For ablation test, four variants of SOLSA were created by disabling filter kernel update, scheduled update and early stop, or a combination of both. By comparing the performance of these variants with SOLSA, we demonstrate the effectiveness of these proposed enhancement techniques in improving the learning of temporal sequences. TABLE II. presents a comparison of different online learning algorithms used in the experiment.

TABLE II. FEATURE OF DIFFERENT ONLINE LEARNING ALGORITHMS

| Feature | SOLSA | E-prop | OTTT | SOLSA variant 1 | SOLSA variant 2 | SOLSA variant 3 |
|---|---|---|---|---|---|---|
| Synapse filter | ✓ |  |  | ✓ | ✓ | ✓ |
| Adaptive weight | ✓ | ✓ | ✓ | ✓ | ✓ | ✓ |
| Adaptive kernel | ✓ |  |  |  | ✓ |  |
| Impact of reset | ✓ | ✓ |  | ✓ | ✓ | ✓ |
| Scheduled updates | ✓ |  |  | ✓ | ✓ | ✓ |
| Early stop | ✓ |  |  |  |  | ✓ |

## A. Performance Comparison

TABLE III. compares different learning algorithms and SOLSA for their accuracy and memory usage. All training uses a batch size of 1 to resemble online learning. The results show that SOLSA achieves the highest accuracy among all algorithms for all datasets. In addition, it requires much lower memory usage compared to BPTT especially when data sequence is long. Please note that, as a classical RNN training algorithm, BPTT calculates the gradient of the weight coefficients by considering their impact to the error of the entire sequence, while SOLSA is only an approximation of BPTT as derived in Section III.B. Theoretically, the performance of SOLSA learning should be slightly inferior to BPTT. However, as we will show later, SOLSA benefits significantly from scheduled weight update and early stop. None of these enhancement techniques are supported by BPTT, because BPTT does not calculate the gradient until the very end of the input sequence. They are only applicable in SOLSA or non-BPTT based algorithm, where gradients are incrementally updated over time.

TABLE III. COMPARISON OF DIFFERENT LEARNING ALGORITHMS

| Dataset | BPTT based | | Three factor Hebbian | | | Memory usage (MB) | |
|---|---|---|---|---|---|---|---|
|  | BPTT | TBPTT | SOLSA | E-prop | OTTT | BPTT | SOLSA |
| EMG gesture | 0.956 | 0.664 | **0.985** | 0.675 | 0.672 | 13.4 | 6.6 |
| Finger mov. | 0.58 | 0.56 | **0.64** | 0.58 | 0.59 | 9.0 | 7.1 |
| Basic motion | 1 | 0.25 | **1** | 0.925 | 1 | 13.4 | 7.5 |
| Epilepsy | 0.941 | 0.676 | **0.971** | 0.816 | 0.904 | 22.7 | 9.9 |
| Jap. Vowel | 0.926 | 0.951 | **0.981** | 0.944 | 0.969 | 7.1 | 6.6 |
| RacketSports | 0.809 | 0.769 | **0.907** | 0.388 | 0.796 | 7.4 | 6.2 |
| DVS128 | 0.959 | 0.6 | **0.979** | 0.819 | 0.875[a] | 105.8 | 67.2 |
| Self reg. scp | 0.836 | 0.866 | **0.897** | 0.876 | 0.89 | 79.6 | 22.2 |
| EMG action | 0.973 | 0.696 | **0.979** | 0.77 | 0.161 | 158.4 | 44.1 |

[a]. Online training using batch 1.

Truncated BPTT effectively reduces the overhead of temporal backpropagation and lowers the memory cost compared with original BPTT, however, as shown in TABLE III. , its performance is unstable. Although it achieves higher accuracy than traditional BPTT on the JapaneseVowels dataset, it performs poorly on other datasets due to its disregard for long-term dependencies in long sequences. Overall, TBPTT's performance is not comparable to that of BPTT and SOLSA.

From TABLE III. , it can be observed that the SOLSA learning algorithm outperforms both E-prop and OTTT algorithms as well. On average, it improves the classification accuracy of the 9 datasets by 30% and 66.5% respectively, compared to E-prop and OTTT. It is worth noting that although the OTTT is a simpler algorithm than E-prop, it updates weights at every time step, while E-prop only updates at the end of the sequence. As a result, OTTT outperforms E-prop for certain datasets.

## B. Ablation Study of SOLSA

In this section, we demonstrate the impact of the enhancement techniques used by SOLSA.

TABLE IV. ABLATION STUDY: SCHEDULED WEIGHT UPDATE

| Dataset | Accuracy | | | |
|---|---|---|---|---|
|  | Unscheduled | Random | Always-on | Scheduled |
| EMG gesture | 0.912 | 0.894 | 0.71 | **0.942** |
| Finger mov. | 0.56 | 0.51 | 0.59 | **0.65** |
| Basic motion | 0.95 | 1/0.5 | 1 | **1** |
| Epilepsy | 0.794 | 0.912 | 0.86 | **0.934** |
| Jap. Vowel | 0.869 | 0.97 | 0.961 | **0.975** |
| RacketSports | 0.598 | 0.828 | 0.756 | **0.855** |
| DVS128 | 0.895 | 0.92 | **0.935** | 0.93 |
| Self reg. scp | 0.88 | 0.89 | 0.88 | **0.894** |
| EMG action | 0.134 | 0.66 | 0.6875 | **0.946** |

We first aim to assess the impact of scheduled weight update on the learning performance. Two heuristic weight update schedules, namely random SOLSA and always-on SOLSA, are created. The random SOLSA employs the same number of update points as the scheduled SOLSA (variant 1), but these update points are randomly selected at different time steps. The always-on SOLSA updates the network at every time step. Additionally, an unscheduled version of SOLSA is also implemented, which updates weights and filter kernels only at the end of sequence. TABLE IV. provides a comparison of these four different update schedules. It shows that, without an update schedule or with heuristic update schedules, SOLSA performs poorly. Comparing the original SOLSA and the version with heuristic update schedules, we can see that the update point selection algorithm leads to approximately 10% accuracy improvement.

We also notice that the accuracy of the EMG action improves from 0.134 to 0.946 after applying scheduled update. Even with a heuristic schedule, the performance improves from 0.134 to 0.66. This indicates that, for very long sequences, updating the model multiple times using the partial gradient obtained from different parts of the sequence is preferable over waiting until the end. Our hypothesis is that updating the model using partial gradient may allow different parts of the network to diverge and specialize in local features within specific portions of the temporal sequence. In data sets like EMG action, the beginning of the time series may contain valuable features that enhance the network's classification performance.

The random SOLSA exhibits unstable performance due to its stochastic nature. It sometimes selects a set of favorable update points and achieves similar performance to the original SOLSA. However, occasionally it chooses a set of extremely poor update points, resulting only 50% of classification accuracy. This demonstrates that in addition to frequent updates, the correct timing of update is also crucial. Simply increasing the number of update points does not necessarily lead to better performance. Updating the network at the right times holds greater importance.

Next, we will show how Adaptive Synapse Filter Kernel and early stop may influence the performance of learning by comparing SOLSA with the 3 variants listed in TABLE II. . For SOLSA with fixed filter kernel, after trying different combinations of $\alpha$ and $\beta$, we set both to be 0.9, which works the best for most of the datasets. TABLE V. presents the comparison results. We can see that the original SOLSA outperforms the variants for almost all cases.

More specifically, the comparison between SOLSA and variant 3 indicates that using adaptive filter kernel helps to improve more than 5% accuracy. We found that, without adapting the filter kernels, the learning performance of variants 1 and 3 are very sensitive to the values of $\alpha$ and $\beta$. We tested different random combinations of $\alpha$ and $\beta$ values on each dataset and recorded the best, worst and medium accuracy of trained models as shown in Fig. 7. Extensive hyperparameter tunning was performed to search for a good combination of kernel coefficients. Eventually we found that setting both $\alpha$ and $\beta$ to 0.9 gives relatively good results for all test cases. The corresponding model accuracies under this setting are given in Fig. 7 in red. And this is the setting used to generate the data in TABLE V. We need to point out that such hyperparameter tuning is usually not practical for online learnings.

Comparing SOLSA with variant 2, we can see that using early stop in training leads to an improvement of 17% in terms of accuracy of the trained model. Using the Epilepsy dataset as an example, Fig. 8 shows the average location where SOLSA stops processing a training sequence over different epoch. At the beginning, the algorithm always trains on the entire sequence. As the training goes on and the model becomes more and more accurate, the early-stop mechanism gets exercised more frequently and a shorter portion of the sequences is being used for the training in average.

TABLE V. ABLATION STUDY: Early stop and adaptive kernel

| Dataset | Accuracy | | | |
|---|---|---|---|---|
| | variant 1 | variant 2 | variant 3 | SOLSA |
| EMG gesture | 0.942 | 0.671 | 0.957 | **0.985** |
| Finger mov. | **0.65** | 0.59 | 0.58 | 0.64 |
| Basic motion | 1 | 1 | 1 | 1 |
| Epilepsy | 0.934 | 0.713 | 0.958 | **0.971** |
| Jap. Vowel | 0.975 | 0.619 | 0.96 | **0.981** |
| RacketSports | 0.855 | 0.901 | 0.835 | **0.907** |
| DVS128 | 0.93 | 0.959 | 0.93 | **0.979** |
| Self reg. scp | 0.894 | 0.897 | 0.893 | **0.897** |
| EMG action | 0.93 | 0.946 | 0.848 | **0.979** |

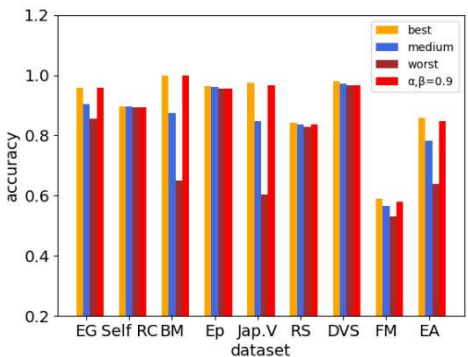

Fig. 7. Accuracies of different $\alpha$ and $\beta$ settings.

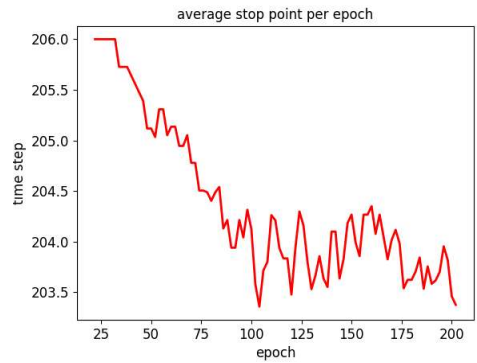

Fig. 8. Average training stop point.

## V. CONCLUSION

In this paper, we present SOLSA, an online learning algorithm for SNNs with both synapse and neuron dynamics. Enhancement techniques such as adaptive filter kernel, scheduled weight update and early stop are also presented. Compared with BPTT training, SOLSA requires much less memory storage and has more balanced temporal workload distribution, hence more suitable for edge implementation. Compared with other online learning algorithms, SOLSA leads to more accurate models with more robust performance. The algorithm is evaluated using data series collected from sensor readings, and the results indicate SOLSA provides outstanding performance in learning temporal sequences compared to existing algorithms.